\documentclass[conference]{IEEEtran}
\IEEEoverridecommandlockouts
\usepackage{cite}
\usepackage{amsmath,amssymb,amsfonts}
\usepackage{algorithmic}
\usepackage{graphicx}
\usepackage{subfigure}
\usepackage{textcomp}
\usepackage{xcolor}
\def\BibTeX{{\rm B\kern-.05em{\sc i\kern-.025em b}\kern-.08em
    T\kern-.1667em\lower.7ex\hbox{E}\kern-.125emX}}
\begin{document}

\title{TAFNet: A Three-Stream Adaptive Fusion Network for RGB-T Crowd Counting\\}

\author{\IEEEauthorblockN{Haihan Tang \quad\qquad \qquad\qquad Yi Wang \quad\qquad\qquad \qquad Lap-Pui Chau}
\IEEEauthorblockA{\textit{School of Electrical and Electronic Engineering, Nanyang Technological University, Singapore} \\
e200249@e.ntu.edu.sg \qquad\qquad wang1241@e.ntu.edu.sg \qquad\qquad  elpchau@ntu.edu.sg}
}

\maketitle

\begin{abstract}
In this paper, we propose a three-stream adaptive fusion network named TAFNet, which uses paired RGB and thermal images for crowd counting. Specifically, TAFNet is divided into one main stream and two auxiliary streams. We combine a pair of RGB and thermal images to constitute the input of main stream. Two auxiliary streams respectively exploit RGB image and thermal image to extract modality-specific features. Besides, we propose an Information Improvement Module (IIM) to fuse the modality-specific features into the main stream adaptively. Experiment results on RGBT-CC dataset show that our method achieves more than 20\% improvement on mean average error and root mean squared error compared with state-of-the-art method. The source code will be publicly available at https://github.com/TANGHAIHAN/TAFNet.

\end{abstract}


\section{Introduction}
As convolutional neural network (CNN) becomes more and more popular in recent years, crowd counting\cite{b10}\cite{b12}
task attracts many researchers' interests. Crowd counting has a variety of applications, such as pedestrian flow monitor \cite{b11} and crowd analysis\cite{b26}.

Before CNN became popular, crowd counting approaches were mainly divided into two groups: counting-by-detection\cite{b15}\cite{b16} and counting-by-regression\cite{b13}\cite{b14}. In recent years, more and more CNN-based methods were proposed, like YOLO\cite{b20} and MCNN\cite{b21}. However, these methods are hard to handle the condition while dense crowds are in high overlap and occlusion. To solve this problem, many approaches based on estimating density map were proposed\cite{b7}\cite{b9}.

\begin{figure}
\centering
\subfigure[]{\includegraphics[width=2.85cm]{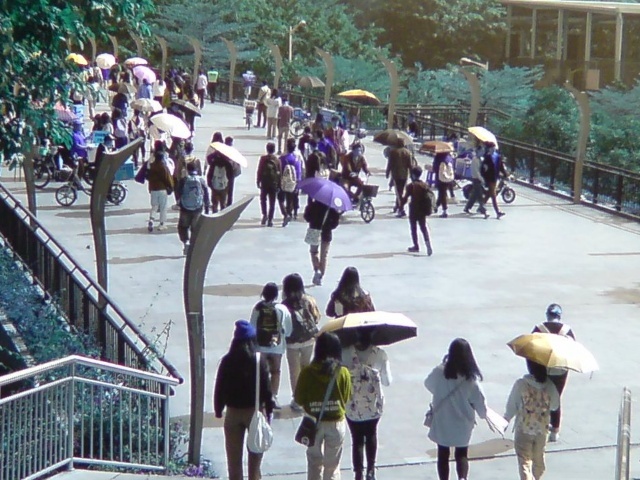}} 
\subfigure[]{\includegraphics[width=2.85cm]{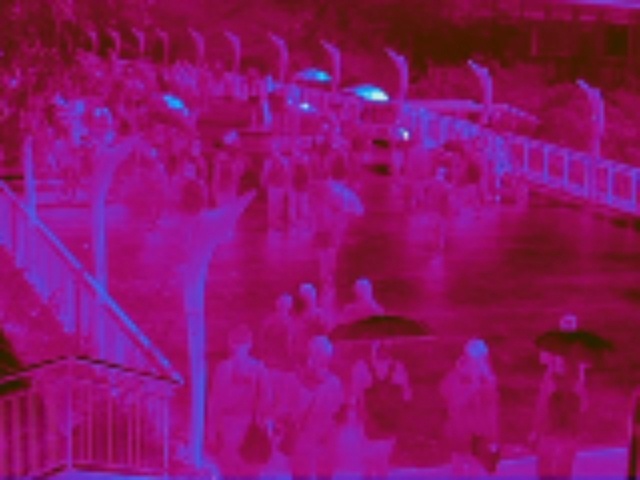}}
\subfigure[]{\includegraphics[width=2.85cm]{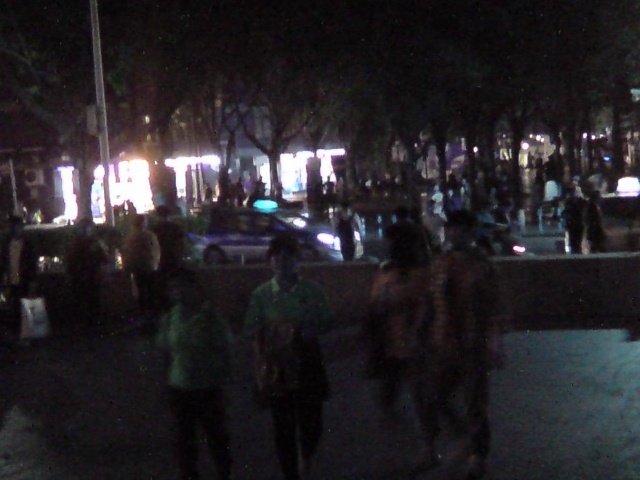}} 
\subfigure[]{\includegraphics[width=2.85cm]{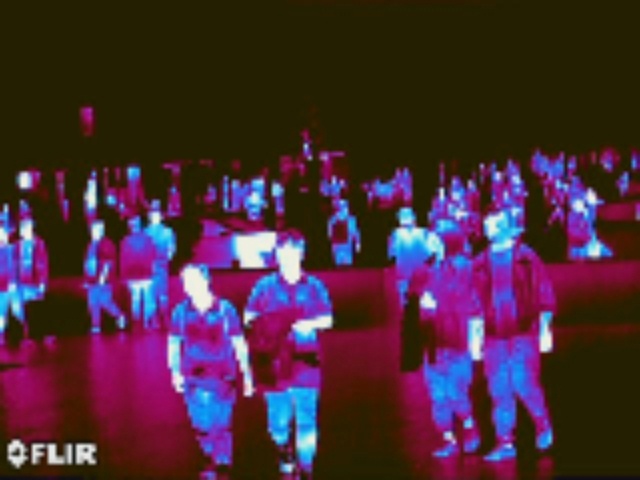}}
\subfigure[]{\includegraphics[width=2.85cm]{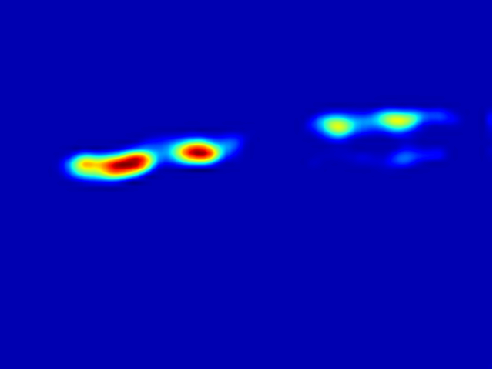}}
\subfigure[]{\includegraphics[width=2.85cm]{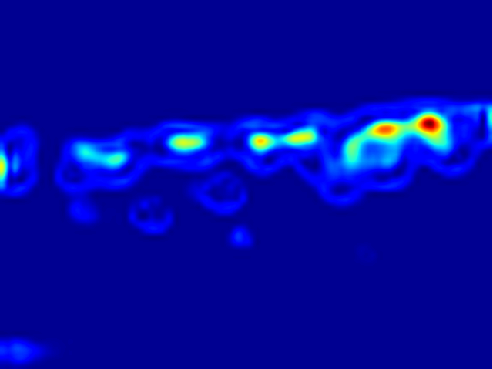}}
\caption{(a)(b): a pair of RGB and thermal images in bright illumination. (c)(d): a pair of RGB and thermal images in dark environment. (e): Estimation result only using RGB image (c) as input. (f): Estimation result using our fusing method to fuse RGB and thermal images (c) and (d) as input.} 
\label{fig3}
\end{figure}

Density prediction methods are based on RGB images to extract features and predict density map. One of the drawbacks is that error will be serious in dark environments where pedestrians are almost invisible. Using optical information and thermal information collaboratively has become one of the solutions. 
Cross-modal learning, i.e., RGB-Thermal (RGB-T) has attracted much attention recently.
Some RGB-T datasets and methods were proposed mainly to address RGB-T tracking\cite{b24}\cite{b25} and RGB-T saliency detection\cite{b22}\cite{b23}. More recently, several RGB-T crowd counting datasets and approaches have been proposed. As shown in Fig.~\ref{fig3}, in bright illumination, RGB image can provide strong evidence while thermal image is hard to distinguish between people and background. On the contrary, thermal image is more clear in the dark environment while RGB image is almost invisible. Therefore, how to fuse RGB and thermal information becomes the main problem.
Liu et al.\cite{b3} proposed a cross-modal collaborative representation learning framework and RGBT-CC dataset. Peng et al.\cite{b17} proposed a novel end-to-end pipeline: MMCC, and DroneRGBT benchmark.
Their methods both use two-input and fuse features extracted from the optical image and thermal image during process. We argue that some detailed information might be lost during fusion. Besides, we find that in RGB-T crowd counting, the estimation results in bright illumination are comparably worse than that in dark environments, which is opposite compared with methods only using RGB images to predict density map. We argue that a more balanced fusing method could improve the estimation result in both bright and dark environments.

\begin{figure*}[t]
\begin{center}
\includegraphics[width=18cm,height=5.14cm]{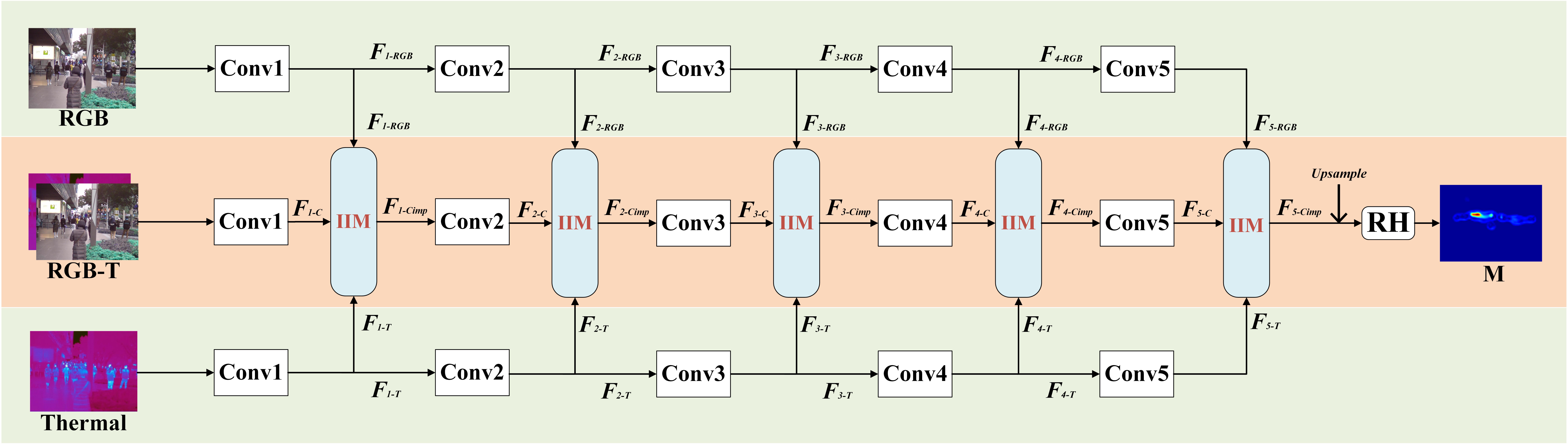}
\caption{Illustration of proposed three-stream adaptive fusion network. `Conv1-5' represents 5 stages of VGG16. After Conv5 in the main stream, we upsample the output, following a Regression Header (RH) to predict the density map M.}
\label{fig1}    
\end{center}
\end{figure*}

In this paper, we proposed a three-stream adaptive fusion network with one main stream and two auxiliary streams. More specifically, the main stream is used to extract combination features, i.e., concatenate a pair of RGB and thermal images and feed them into main stream to extract combination features. Two auxiliary streams extract modality-specific features from RGB and thermal images, respectively. The Information Improvement Module (IIM) is proposed to adaptively fuse the combination features and modality-specific features, improving feature representation of main stream. 
Compared with other methods, our method can retain the features to a better extent.

Experimental results show that our method outperforms state-of-the-art RGBT method to a great extent, i.e., more than 20\% on the root mean square error (RMSE) and mean average error (MAE). Further ablation studies verify the effectiveness of our proposed three-stream architecture and Information Improvement Module. In summary, our contributions are as follows:
\begin{itemize}
\item Different from two-input naive architecture for two-modality data, we proposed a three-input network for RGB-T crowd counting. Three streams in our network are fused by Information Improvement Module to predict density map. As far as we know, we are the first one to propose three-input network in RGB-T crowd counting.

\item We proposed an Information Improvement Module (IIM) to improve feature representation of the main stream by fusing combination features with modality-specific features adaptively. The IIM could effectively use combination features and modality-specific features, at the same time, reduce redundant information.

\item Previous single modality methods have a large estimation error in dark environments, while cross-modal methods have an inaccurate estimate in bright illumination. Our network could reach an accurate estimation result in both bright and dark environments. 
\end{itemize}

\section{PROPOSED METHOD}
We proposed a three-stream adaptive fusion network for RGB-T crowd counting and Information Improvement Module (IIM) to enhance the feature representation. In this part, the architecture of our network will be introduced first. Following is the description of IIM.

\subsection{Network Architecture}\label{AA}
As shown in Fig.~\ref{fig1}, our proposed architecture has three streams, including one main stream in the middle row (with orange background) and two auxiliary streams in the top and bottom rows (with green background). Inspired by \cite{b1}, we concatenate paired RGB and thermal images and feed them into main stream to extract combination features. Different from them, since the fusion process requires to fuse features extracted from the same stage, we use three VGG16\cite{b2} networks as the backbone. At the same time, RGB and thermal images are separately fed into two auxiliary streams to extract modality-specific features. Three streams are connected by Information Improvement Module (IIM). 

More specifically, in the main stream, we concatenate paired RGB and thermal images and feed them into a VGG16 network. Here we modify the input layer of VGG16 in order to satisfy the concatenate image input. After each stage, we get the combination feature, namely $F_{i-C}$ (i=1,2,3,4,5). At the same time, in two auxiliary streams, paired RGB and thermal images are separately fed into two standard VGG16 networks and get modality-specific features $F_{i-T}$ and $F_{i-RGB}$ (i=1,2,3,4,5) after each stage. We apply IIM after each stage to fuse $F_{i-T}$, $F_{i-RGB}$ and $F_{i-C}$. This process can be formulated as:
\begin{equation}
F_{i-C} \emph{imp} =IIM(F_{i-T}, F_{i-RGB},F_{i-C})\label{eq1}
\end{equation}
where $F_{i-C}$\emph{imp} is the improved combination feature. In this way, we can improve the combination feature with modality-specific features. The improved combination feature will then be directly fed into the next stage of the main stream. After stage 5, we upsample the output $F_{5-C}$\emph{imp}, following a regression header to predict the density map M.

\subsection{Information Improvement Module}

In the main stream, the feature is extracted from the concatenate of paired RGB and thermal images. Since the paired RGB and thermal images are captured by different sensors, paired images are inevitably not strictly aligned. For that reason, we believe that combination features lost some detail in RGB and thermal images to some extent. To improve the feature representation of main stream, we propose Information Improvement Module to better fuse the combination feature with modality-specific features. In this way, the detail of RGB and thermal features can be compensated for the combination feature, and the misaligned problem could be solved, which results in a better performance in density map estimation.

\begin{figure}[h]
\centerline{\includegraphics[width=8.8cm,height=4.622cm]{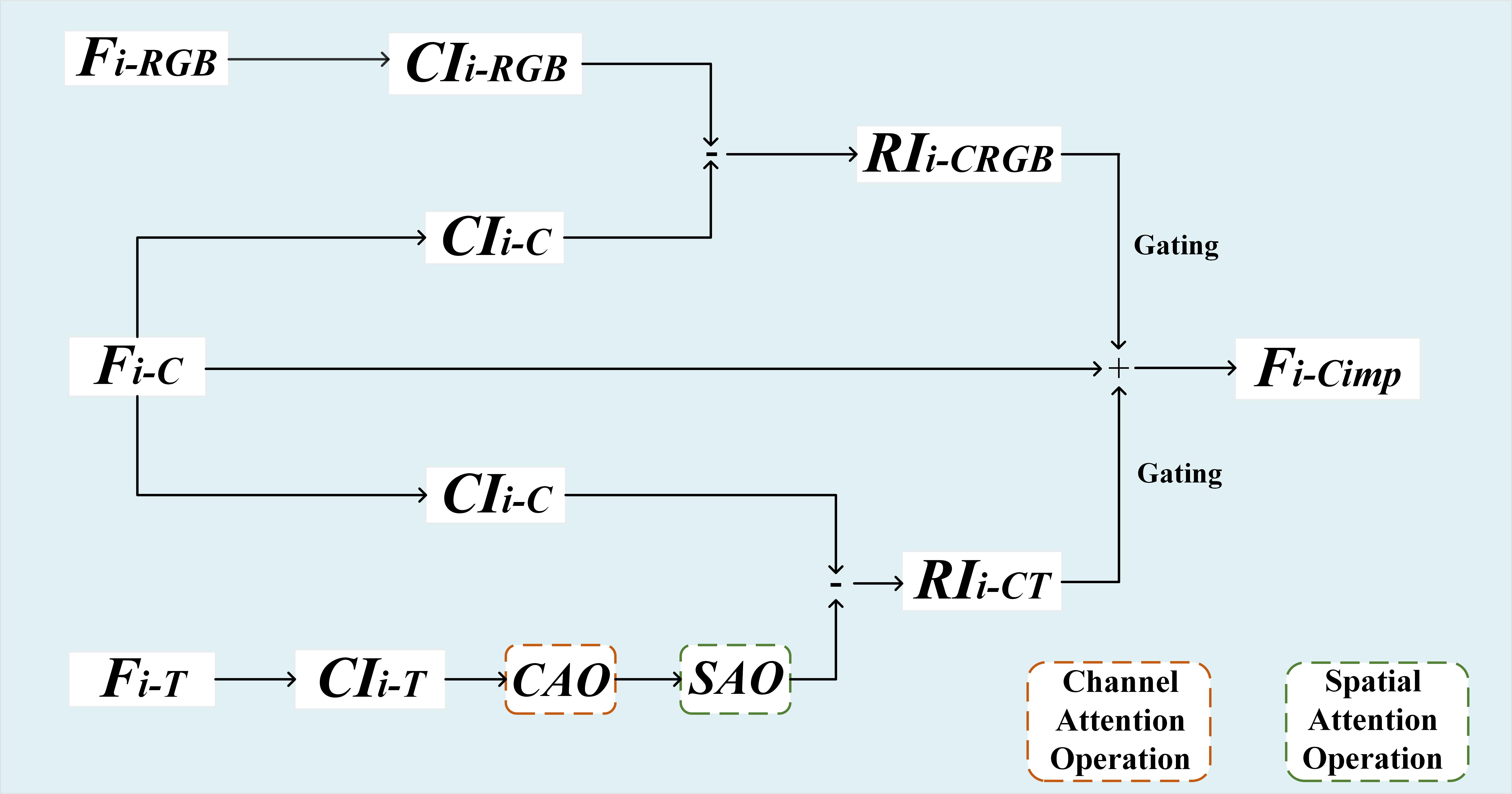}}
\caption{Information Improvement Module}
\label{fig2}
\end{figure}

As shown in Fig.~\ref{fig2}, after we get $F_{i-T}$, $F_{i-RGB}$ and $F_{i-C}$ from $i^{th}$ stage, we apply pyramid pooling layer to extract contextual information $CI_{i-T}$, $CI_{i-RGB}$ and $CI_{i-C}$ respectively, as \cite{b3} did. By using contextual information can we avoid excessive fusing of features and mitigate the misalignment in RGBT-CC dataset.

Since the thermal image can provide strong support on density map estimation, especially in the dark background, we believe that features from thermal image require adaptive refinement. We apply Channel Attention Operation and Spatial Attention Operation\cite{b4}\cite{b5} sequentially to the contextual information $CI_{i-T}$ and then compute the difference with $CI_{i-C}$ to get the residual information $RI_{i-CT}$. Similarly, we compute the difference between $CI_{i-RGB}$ and $CI_{i-C}$ to get the residual information $RI_{i-CRGB}$. Finally, we apply Gating mechanism, i.e., giving weights to two residual information and adding them to combination features, producing $F_{i-Cimp}$.

\section{EXPERIMENTS}

In this part, we will first introduce our experiment details, following the experimental evaluations. Then, the comparison with state-of-the-art methods and the ablation study results will be discussed.

\subsection{Implementation Details}
Our experiment was implemented with PyTorch\cite{b6}. We used BL\cite{b7} as the backbone and built our structure based on\cite{b3}. We used RGBT-CC\cite{b3} dataset which contains 1030, 200, 800 pairs of RGB-T images as training set, validation set and test set, respectively. Max training epoch was set to 300. During the training process, after 20 epochs, we used validation set to verify if it is the best model at each epoch. Then we used test set to test each best model. We used Adam\cite{b8} for optimization. To avoid overfitting, we set weight decay to 1e-4. Learning rate was set to 1e-5. Training process will take about 11 hours on a single NVIDIA GeForce RTX 2080 Ti GPU.

\subsection{Experimental Evaluations}
We use two primary metrics to evaluate our model: Root Mean Square Error (RMSE) and Grid Average Mean Absolute Error (GAME). For both RMSE and GAME, lower value means better performance. RMSE is defined as:
\begin{equation}
RMSE = \sqrt{\frac{1}{n}\Sigma_{i=1}^{n}{\Big({P_i -G_i}\Big)^2}}
\end{equation}
where ${P_i}$ and ${G_i}$ refer to the predicted count and ground truth of ${i^{th}}$ test picture pairs. n is the total number of paired pictures in test set. To calculate GAME, we grid the picture into $4^l$ parts and calculate the counting error in each part, finally adding them together. GAME is defined as:
\begin{equation}
GAME(l) = \frac{1}{n}\Sigma_{i=1}^{n}\Sigma_{j=1}^{4^l}{|{P_i^j -G_i^j}|}
\end{equation}
where ${P_i^j}$ and ${G_i^j}$ refer to the predicted count and ground truth in ${j^{th}}$ part of ${i^{th}}$ test picture pairs. It is worth mentioning that GAME(0) is equivalent to Mean Absolute Error (MAE).

\subsection{Comparison with state-of-the-art methods}
\begin{figure}[htbp]
\centerline{\includegraphics[width=8.8cm,height=7.084cm]{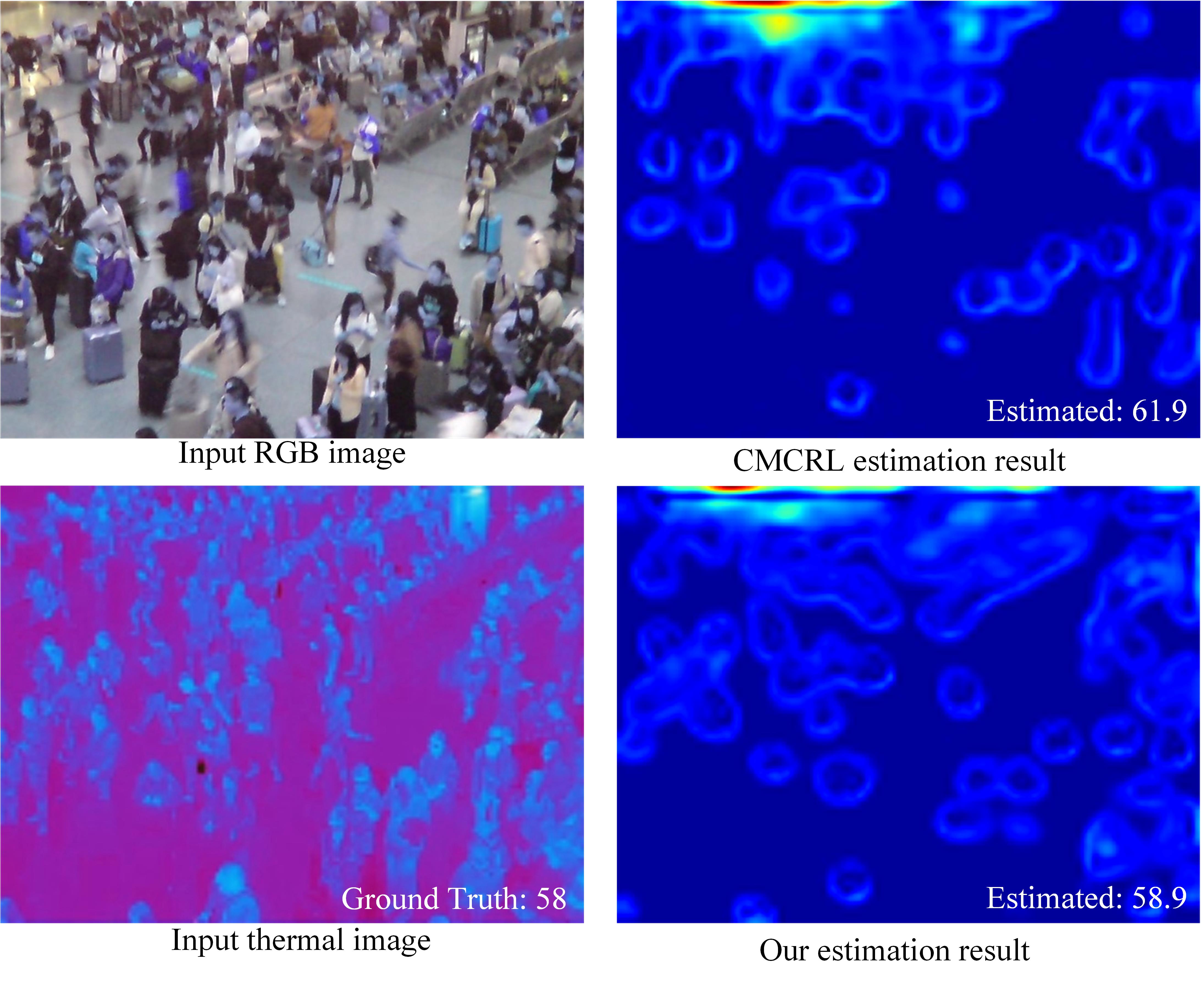}}
\caption{Comparison of our estimated density map with CMCRL[3]}
\label{fig4}
\end{figure}

We compared the proposed method with several state-of-the-art methods, including CSRNet\cite{b9}, BL\cite{b7}, CMCRL\cite{b3}. As shown in Table \ref{table1}, our method outperforms any other SOTA methods. To illustrate in detail, compared with CMCRL who achieves the best result for now, our method achieves 20.7\%, 14.9\%, 11.4\%, 8.2\%, 20.3\% improvement on GAME(0), GAME(1), GAME(2), GAME(3) and RMSE respectively. Comparison of estimated density map is shown in Fig.~\ref{fig4}. Our method could reach a more accurate result, especially in the dense crowd.
We also did experiments to test our model's performance in different illumination conditions. The result is shown in Table \ref{table2}. Compared with CMCRL, our method achieves 23.5\%, 12.4\%, 6.4\%, 0.3\%, 25.5\% improvement on GAME(0), GAME(1), GAME(2), GAME(3) and RMSE respectively in bright condition, while having slightly better results in dark condition. Besides, our method is more stable between bright and dark conditions. In other words, our method can achieve a more accurate prediction in both bright and dark environments.

\begin{table}[htbp]
\caption{Performance of our model compared with other state-of-the-art methods}
\begin{center}
\resizebox{\linewidth}{!}{
\begin{tabular}{|l|c|c|c|c|c|}
\hline
\textbf{Model}  & \multicolumn{1}{l|}{\textit{\textbf{GAME(0)}}} & \multicolumn{1}{l|}{\textit{\textbf{GAME(1)}}} & \multicolumn{1}{l|}{\textit{\textbf{GAME(2)}}} & \multicolumn{1}{l|}{\textit{\textbf{GAME(3)}}} & \multicolumn{1}{l|}{\textit{\textbf{RMSE}}} \\ \hline
\textbf{CSRNet[9]} & 20.40                                          & 23.58                                          & 28.03                                          & 35.51                                          & 35.26                                       \\ \hline
\textbf{BL[7]}     & 18.70                                          & 22.55                                          & 26.83                                          & 34.62                                          & 32.67                                       \\ \hline
\textbf{CMCRL[3]}  & 15.61                                          & 19.95                                          & 24.69                                          & 32.89                                          & 28.18                                       \\ \hline
\textbf{Ours}   & {\textbf{12.38}}          & {\textbf{16.98}}          & {\textbf{21.86}}          & {\textbf{30.19}}          & {\textbf{22.45}}       \\ \hline
\end{tabular}}
\end{center}
\label{table1}
\end{table}

\begin{table}[htbp]
\caption{Performance of our model compared with CMCRL in different illumination conditions on RGBT-CC dataset}
\centering
\resizebox{\linewidth}{!}{
\begin{tabular}{|c|c|c|c|c|c|c|} 
\hline
\textbf{Illumination} & \textbf{Model} & \textbf{\textit{GAME(0)}}       & \textbf{\textit{GAME(1)}}       & \textbf{\textit{GAME(2)}}       & \textbf{\textit{GAME(3)}}       & \textbf{\textit{RMSE}}           \\ 
\hline
\textbf{Bright}       & \textbf{CMCRL[3]} & 20.36                           & 23.57                           & 28.49                           & 36.29                           & 32.57                            \\
                      & \textbf{Ours}  & {\textbf{15.57}} & {\textbf{20.65}} & {\textbf{26.67}} & {\textbf{36.17}} & {\textbf{24.25}}  \\ 
\hline
\textbf{Dark}         & \textbf{CMCRL[3]} & 15.44                           & 19.23                           & 23.79                           & 30.28                           & 29.11                            \\
                      & \textbf{Ours}  & \textbf{{14.20}} & \textbf{{19.20}} & \textbf{{24.00}} & \textbf{{31.63}} & \textbf{{27.50}}  \\
\hline
\end{tabular}}
\label{table2}
\end{table}

\subsection{Ablation Study}
We have done ablation studies on RGBT-CC dataset to test the performance of key modules in our architecture. As shown in Table \ref{table3}, the baseline is constructed without the whole IIM module. In other words, the architecture of baseline is one-stream architecture with concatenate image as input to predict density map. We then add IIM module but without `CAO\&SAO' to the baseline and forms `w. IIM(w/o CAO\&SAO)'. Finally, we add the whole IIM module to baseline, i.e., `w. IIM', which is also our proposed structure. The result shows that both `CAO\&SAO' and IIM can decrease the prediction errors. Having both of them can achieve the best performance.

\begin{table}[htbp]
\caption{Ablation study}
\begin{center}
\resizebox{\linewidth}{!}{
\begin{tabular}{|c|c|c|c|c|c|} 
\hline
\textbf{Model}                & \textbf{\textit{GAME(0)}} & \textbf{\textbf{\textit{GAME(1)}}} & \textbf{\textbf{\textit{GAME(2)}}} & \textbf{\textbf{\textit{GAME(3)}}} & \textbf{\textit{RMSE}}  \\ 
\hline
\textbf{CMCRL[3]}                & 15.61                     & 19.95                              & 24.69                              & 32.89                              & 28.18                   \\ 
\hline
\textbf{baseline}             & 16.34                     & 20.64                              & 25.32                              & 33.39                              & 29.03                   \\ 
\hline
\textbf{w. IIM(w/o CAO\&SAO)} & 14.13                     & 18.37                              & 23.21                              & 31.49                              & 24.40                   \\ 
\hline
\textbf{w.IIM}                & \textbf{12.38}            & \textbf{16.98}                     & \textbf{21.86}                     & \textbf{30.19}                     & \textbf{22.45}          \\
\hline
\end{tabular}}
\end{center}
\label{table3}
\end{table}


 \section{CONCLUSION}
In this paper, we proposed a three-stream architecture for RGB-T crowd counting. The main stream in our architecture can extract rich information, and two auxiliary streams are used to improve the modality-specific features to the main stream continuously. Moreover, the Information Improvement Module was proposed to adaptively fuse combination features with modality-specific features, which further decreases the error in RGB-T crowd counting. Experiment shows our approach outperforms other state-of-the-art methods for RGB-T crowd counting. Ablation studies verified that the three-stream network and IIM we proposed are effective. Besides, our method can reach a more stable result in both bright illumination and dark environments.


\begin{thebibliography}{10}
\providecommand{\url}[1]{#1}
\csname url@samestyle\endcsname
\providecommand{\newblock}{\relax}
\providecommand{\bibinfo}[2]{#2}
\providecommand{\BIBentrySTDinterwordspacing}{\spaceskip=0pt\relax}
\providecommand{\BIBentryALTinterwordstretchfactor}{4}
\providecommand{\BIBentryALTinterwordspacing}{\spaceskip=\fontdimen2\font plus
\BIBentryALTinterwordstretchfactor\fontdimen3\font minus
  \fontdimen4\font\relax}
\providecommand{\BIBforeignlanguage}[2]{{%
\expandafter\ifx\csname l@#1\endcsname\relax
\typeout{** WARNING: IEEEtran.bst: No hyphenation pattern has been}%
\typeout{** loaded for the language `#1'. Using the pattern for}%
\typeout{** the default language instead.}%
\else
\language=\csname l@#1\endcsname
\fi
#2}}
\providecommand{\BIBdecl}{\relax}
\BIBdecl

\bibitem{9466383}
W.~Zhou, S.~Pan, J.~Lei, and L.~Yu, ``Mrinet: Multilevel reverse-context
  interactive-fusion network for detecting salient objects in rgb-d images,''
  \emph{IEEE Signal Processing Letters}, vol.~28, pp. 1525--1529, 2021.

\bibitem{Simonyan2015VeryDC}
K.~Simonyan and A.~Zisserman, ``Very deep convolutional networks for
  large-scale image recognition,'' \emph{CoRR}, vol. abs/1409.1556, 2015.

\bibitem{Liu2021CrossModalCR}
L.~Liu, J.~Chen, H.~Wu, G.~Li, C.~Li, and L.~Lin, ``Cross-modal collaborative
  representation learning and a large-scale rgbt benchmark for crowd
  counting,'' in \emph{CVPR}, 2021.

\bibitem{Woo2018CBAMCB}
S.~Woo, J.~Park, J.-Y. Lee, and I.-S. Kweon, ``Cbam: Convolutional block
  attention module,'' in \emph{ECCV}, 2018.

\bibitem{Fan2020BBSNetRS}
D.-P. Fan, Y.~Zhai, A.~Borji, J.~Yang, and L.~Shao, ``Bbs-net: Rgb-d salient
  object detection with a bifurcated backbone strategy network,'' \emph{ArXiv},
  vol. abs/2007.02713, 2020.

\bibitem{Paszke2019PyTorchAI}
A.~Paszke, S.~Gross, F.~Massa, A.~Lerer, J.~Bradbury, G.~Chanan, T.~Killeen,
  Z.~Lin, N.~Gimelshein, L.~Antiga, A.~Desmaison, A.~K{\"o}pf, E.~Yang,
  Z.~DeVito, M.~Raison, A.~Tejani, S.~Chilamkurthy, B.~Steiner, L.~Fang,
  J.~Bai, and S.~Chintala, ``Pytorch: An imperative style, high-performance
  deep learning library,'' \emph{ArXiv}, vol. abs/1912.01703, 2019.

\bibitem{Ma2019BayesianLF}
Z.~Ma, X.~Wei, X.~Hong, and Y.~Gong, ``Bayesian loss for crowd count estimation
  with point supervision,'' \emph{2019 IEEE/CVF International Conference on
  Computer Vision (ICCV)}, pp. 6141--6150, 2019.

\bibitem{Kingma2015AdamAM}
D.~P. Kingma and J.~Ba, ``Adam: A method for stochastic optimization,''
  \emph{CoRR}, vol. abs/1412.6980, 2015.

\bibitem{Li2018CSRNetDC}
Y.~Li, X.~Zhang, and D.~Chen, ``Csrnet: Dilated convolutional neural networks
  for understanding the highly congested scenes,'' \emph{2018 IEEE/CVF
  Conference on Computer Vision and Pattern Recognition}, pp. 1091--1100, 2018.

\bibitem{Gao2020CNNbasedDE}
G.~Gao, J.~Gao, Q.~Liu, Q.~Wang, and Y.~Wang, ``Cnn-based density estimation
  and crowd counting: A survey,'' \emph{ArXiv}, vol. abs/2003.12783, 2020.

\bibitem{Sutheerakul2017ApplicationOU}
C.~Sutheerakul, N.~Kronprasert, M.~Kaewmoracharoen, and P.~Pichayapan,
  ``Application of unmanned aerial vehicles to pedestrian traffic monitoring
  and management for shopping streets,'' \emph{Transportation research
  procedia}, vol.~25, pp. 1717--1734, 2017.

\bibitem{Boominathan2016CrowdNetAD}
L.~Boominathan, S.~S.~S. Kruthiventi, and R.~V. Babu, ``Crowdnet: A deep
  convolutional network for dense crowd counting,'' \emph{Proceedings of the
  24th ACM international conference on Multimedia}, 2016.

\bibitem{Idrees2013MultisourceMC}
H.~Idrees, I.~Saleemi, C.~Seibert, and M.~Shah, ``Multi-source multi-scale
  counting in extremely dense crowd images,'' \emph{2013 IEEE Conference on
  Computer Vision and Pattern Recognition}, pp. 2547--2554, 2013.

\bibitem{Wang2019ObjectCI}
Y.~Wang, J.~Hou, and L.-P. Chau, ``Object counting in video surveillance using
  multi-scale density map regression,'' \emph{ICASSP 2019 - 2019 IEEE
  International Conference on Acoustics, Speech and Signal Processing
  (ICASSP)}, pp. 2422--2426, 2019.

\bibitem{Liu2019HighLevelSF}
W.~Liu, S.~Liao, W.~Ren, W.~Hu, and Y.~Yu, ``High-level semantic feature
  detection: A new perspective for pedestrian detection,'' \emph{2019 IEEE/CVF
  Conference on Computer Vision and Pattern Recognition (CVPR)}, pp.
  5182--5191, 2019.

\bibitem{Wang2021ASA}
Y.~Wang, J.~Hou, X.~Hou, and L.-P. Chau, ``A self-training approach for
  point-supervised object detection and counting in crowds,'' \emph{IEEE
  Transactions on Image Processing}, vol.~30, pp. 2876--2887, 2021.

\bibitem{Peng2020RGBTCC}
T.~Peng, Q.~Li, and P.~Zhu, ``Rgb-t crowd counting from drone: A benchmark and
  mmccn network,'' in \emph{ACCV}, 2020.

\bibitem{Topkaya2014CountingPB}
I.~S. Topkaya, H.~Erdogan, and F.~M. Porikli, ``Counting people by clustering
  person detector outputs,'' \emph{2014 11th IEEE International Conference on
  Advanced Video and Signal Based Surveillance (AVSS)}, pp. 313--318, 2014.

\bibitem{Redmon2016YouOL}
J.~Redmon, S.~K. Divvala, R.~B. Girshick, and A.~Farhadi, ``You only look once:
  Unified, real-time object detection,'' \emph{2016 IEEE Conference on Computer
  Vision and Pattern Recognition (CVPR)}, pp. 779--788, 2016.

\bibitem{Zhang2016SingleImageCC}
Y.~Zhang, D.~Zhou, S.~Chen, S.~Gao, and Y.~Ma, ``Single-image crowd counting
  via multi-column convolutional neural network,'' \emph{2016 IEEE Conference
  on Computer Vision and Pattern Recognition (CVPR)}, pp. 589--597, 2016.

\bibitem{Zhang2020DSiamMFTAR}
X.~Zhang, P.~Ye, S.~Peng, J.~Liu, and G.~Xiao, ``Dsiammft: An rgb-t fusion
  tracking method via dynamic siamese networks using multi-layer feature
  fusion,'' \emph{Signal Process. Image Commun.}, vol.~84, p. 115756, 2020.

\bibitem{Yang2019LearningTD}
R.~Yang, Y.~Zhu, X.~Wang, C.~Li, and J.~Tang, ``Learning target-oriented dual
  attention for robust rgb-t tracking,'' \emph{2019 IEEE International
  Conference on Image Processing (ICIP)}, pp. 3975--3979, 2019.

\bibitem{Li2018AUR}
C.~Li, G.~Wang, Y.~Ma, A.~Zheng, B.~Luo, and J.~Tang, ``A unified rgb-t
  saliency detection benchmark: Dataset, baselines, analysis and a novel
  approach,'' \emph{ArXiv}, vol. abs/1701.02829, 2018.

\bibitem{Tang2020RGBTSO}
J.~Tang, D.~Fan, X.~Wang, Z.~Tu, and C.~Li, ``Rgbt salient object detection:
  Benchmark and a novel cooperative ranking approach,'' \emph{IEEE Transactions
  on Circuits and Systems for Video Technology}, vol.~30, pp. 4421--4433, 2020.

\bibitem{Li2015CrowdedSA}
T.~Li, H.~Chang, M.~Wang, B.~Ni, R.~Hong, and S.~Yan, ``Crowded scene analysis:
  A survey,'' \emph{IEEE Transactions on Circuits and Systems for Video
  Technology}, vol.~25, pp. 367--386, 2015.

\end{thebibliography}


\begin{thebibliography}{00}
\bibitem{b1} W. Zhou, S. Pan, J. Lei, and L. Yu, “Mrinet: Multilevel reverse-context
interactive-fusion network for detecting salient objects in rgb-d images,”
IEEE Signal Processing Letters, vol. 28, pp. 1525–1529, 2021.
\bibitem{b2} K. Simonyan and A. Zisserman, “Very deep convolutional networks for large-scale image recognition,” CoRR, vol. abs/1409.1556, 2015.
\bibitem{b3}  L. Liu, J. Chen, H. Wu, G. Li, C. Li, and L. Lin, “Cross-modal collaborative representation learning and a large-scale rgbt benchmark for crowd
counting,” in CVPR, 2021.
\bibitem{b4} S. Woo, J. Park, J.-Y. Lee, and I.-S. Kweon, “Cbam: Convolutional block
attention module,” in ECCV, 2018.
\bibitem{b5}  D.-P. Fan, Y. Zhai, A. Borji, J. Yang, and L. Shao, “Bbs-net: Rgb-d salient
object detection with a bifurcated backbone strategy network,” ArXiv, vol.
abs/2007.02713, 2020.
\bibitem{b6} A. Paszke, S. Gross, F. Massa, A. Lerer, J. Bradbury, G. Chanan,
T. Killeen, Z. Lin, N. Gimelshein, L. Antiga, A. Desmaison, A. K¨opf,
E. Yang, Z. DeVito, M. Raison, A. Tejani, S. Chilamkurthy, B. Steiner,
L. Fang, J. Bai, and S. Chintala, “Pytorch: An imperative style, highperformance deep learning library,” ArXiv, vol. abs/1912.01703, 2019.
\bibitem{b7}  Z. Ma, X. Wei, X. Hong, and Y. Gong, “Bayesian loss for crowd count estimation with point supervision,” 2019 IEEE/CVF International Conference
on Computer Vision (ICCV), pp. 6141–6150, 2019.
\bibitem{b8} D. P. Kingma and J. Ba, “Adam: A method for stochastic optimization,”
CoRR, vol. abs/1412.6980, 2015.
\bibitem{b9}  Y. Li, X. Zhang, and D. Chen, “Csrnet: Dilated convolutional neural networks for understanding the highly congested scenes,” 2018 IEEE/CVF
Conference on Computer Vision and Pattern Recognition, pp. 1091–1100,
2018.

\bibitem{b10} G. Gao, J. Gao, Q. Liu, Q. Wang, and Y. Wang, “Cnn-based density
estimation and crowd counting: A survey,” ArXiv, vol. abs/2003.12783,
2020.
\bibitem{b11} C. Sutheerakul, N. Kronprasert, M. Kaewmoracharoen, and P. Pichayapan,
“Application of unmanned aerial vehicles to pedestrian traffic monitoring
and management for shopping streets,” Transportation research procedia,
vol. 25, pp. 1717–1734, 2017.
\bibitem{b12} L. Boominathan, S. S. S. Kruthiventi, and R. V. Babu, “Crowdnet: A deep convolutional network for dense crowd counting,” Proceedings of the 24th ACM international conference on Multimedia, 2016.
\bibitem{b13} H. Idrees, I. Saleemi, C. Seibert, and M. Shah, “Multi-source multi-scale
counting in extremely dense crowd images,” 2013 IEEE Conference on
Computer Vision and Pattern Recognition, pp. 2547–2554, 2013.
\bibitem{b14} Y. Wang, J. Hou, and L.-P. Chau, “Object counting in video surveillance using multi-scale density map regression,” ICASSP 2019 - 2019
IEEE International Conference on Acoustics, Speech and Signal Processing
(ICASSP), pp. 2422–2426, 2019.
\bibitem{b15} W. Liu, S. Liao, W. Ren, W. Hu, and Y. Yu, “High-level semantic feature
detection: A new perspective for pedestrian detection,” 2019 IEEE/CVF
Conference on Computer Vision and Pattern Recognition (CVPR), pp.
5182–5191, 2019.
\bibitem{b16}  Y. Wang, J. Hou, X. Hou, and L.-P. Chau, “A self-training approach for
point-supervised object detection and counting in crowds,” IEEE Transactions on Image Processing, vol. 30, pp. 2876–2887, 2021.
\bibitem{b17} T. Peng, Q. Li, and P. Zhu, “Rgb-t crowd counting from drone: A benchmark and mmccn network,” in ACCV, 2020.
\bibitem{b20} J. Redmon, S. K. Divvala, R. B. Girshick, and A. Farhadi, “You only
look once: Unified, real-time object detection,” 2016 IEEE Conference on
Computer Vision and Pattern Recognition (CVPR), pp. 779–788, 2016.

\bibitem{b21} Y. Zhang, D. Zhou, S. Chen, S. Gao, and Y. Ma, “Single-image crowd
counting via multi-column convolutional neural network,” 2016 IEEE Conference on Computer Vision and Pattern Recognition (CVPR), pp. 589–597, 2016.
\bibitem{b24} X. Zhang, P. Ye, S. Peng, J. Liu, and G. Xiao, “Dsiammft: An rgb-t fusion
tracking method via dynamic siamese networks using multi-layer feature
fusion,” Signal Process. Image Commun., vol. 84, p. 115756, 2020.
\bibitem{b25} R. Yang, Y. Zhu, X. Wang, C. Li, and J. Tang, “Learning target-oriented
dual attention for robust rgb-t tracking,” 2019 IEEE International Conference on Image Processing (ICIP), pp. 3975–3979, 2019.
\bibitem{b22}  C. Li, G. Wang, Y. Ma, A. Zheng, B. Luo, and J. Tang, “A unified rgbt saliency detection benchmark: Dataset, baselines, analysis and a novel
approach,” ArXiv, vol. abs/1701.02829, 2018.
\bibitem{b23} J. Tang, D. Fan, X. Wang, Z. Tu, and C. Li, “Rgbt salient object detection:
Benchmark and a novel cooperative ranking approach,” IEEE Transactions
on Circuits and Systems for Video Technology, vol. 30, pp. 4421–4433, 2020.
\bibitem{b26} T. Li, H. Chang, M. Wang, B. Ni, R. Hong, and S. Yan, “Crowded sceneanalysis:  A survey,” IEEE Transactions on Circuits and Systems for VideoTechnology, vol. 25, pp. 367–386, 2015.

\end{thebibliography}
\end{document}